# Vibrotactile Feedback for a Remote Operated Robot with Noise Subtraction Based on Perceived Intensity

Ryoma Yamawaki, Takeru Shimamura, Noel Avila, Masashi Konyo, Shotaro Kojima, Ranulfo Bezerra, and Satoshi Tadokoro

[1] Graduate School of Information Sciences, Tohoku University, Sendai, Japan

(Email: konyo@rm.is.tohoku.ac.jp)

**Abstract** --- There is a growing demand for tele-operated robots. This paper presents a novel method for reducing vibration noise generated by robot's own motion, which can disrupt the quality of tactile feedback for tele-operated robots. Our approach focuses on perceived intensity, the amount of how humans experience vibration, to create a noise filter that aligns with human perceptual characteristics. This system effectively subtracts ego-noise while preserving the essential tactile signals, ensuring more accurate and reliable haptic feedback for operators. This method offers a refined solution to the challenge of maintaining high-quality tactile feedback in tele-operated systems.

## 1 INTRODUCTION

In recent years, tele-operated robots have played an important role in many areas. To increase the operability of the tele-operated robots, there has been an approach to feedback vibration signals that propagate to the robot chassis [1]. Vibration presentation based on vibration sensing has been found to improve operator operability [2]. However, ego-noise generated by their own motors due to their own motion can adversely affect the operator's perception and reduce the accuracy and efficiency of operation if the vibration of the robot chassis is transmitted as it is.

Therefore, reducing ego-noise is an important issue for improving the quality of vibration feedback to sensing. Conventional noise reduction methods include signal filtering. Since the generated noise may be in the same frequency range as the vibration information to be acquired, it is difficult to remove noise using high-pass or low-pass filters. Therefore, a method that calculates the amount of removal for each frequency based on the pre-recorded ego-noise has been proposed [3].

However, amplitude-based subtraction does not consider the frequency characteristics of human perception, making intensity tuning between stimuli with different frequencies problematic.

The purpose of this paper is to propose a noise filter that takes human perceptual characteristics into account by focusing on the perceived intensity of vibration information. Perceived intensity is a psychophysical quantity that represents the intensity of the vibration stimulus perceived, as defined by Bensmia et al [4].

## 2 METHOD

### 2.1 Intensity Segment Modulation

The authors propose Intensity Segment Modulation (ISM) as a signal processing that converts vibration into an amplitude-modulated wave of any carrier frequency while maintaining the tactile sensation of the vibration [5]. ISM decomposes the input waveform into frequency-based basis signals using Empirical Mode Decomposition (EMD), calculates the perceptual intensity, and determines its total sum. This enables the conversion of high-frequency vibrations above 100 Hz into amplitude-modulated waves of any frequency while maintaining the perceptual quantity.

This method resolves issues related to audible noise and vibrator response limitations, which were challenges in conventional vibration feedback. In this study, the frequency was set to 200 Hz, which is easily perceived by humans. The definition of perceptual intensity used in this paper is shown in equation (1). $A$ is an amplitude of vibration signal. $AT(f)$ and $\alpha(f)$ are coefficients related to the vibration discrimination threshold and the exponent for calculating vibration intensity, respectively, and depend on the frequency $f$.

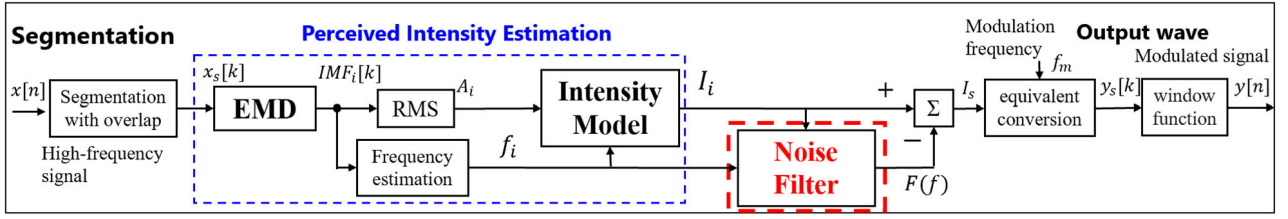

Fig.1 Block Diagram of ISM

$$I(f) = \left[\left(\frac{A}{AT(f)}\right)^2\right]^{\alpha(f)} \quad (1)$$

## 2.2 Perception Based Noise Filter

As a method for removing ego noise generated by robots, we propose a noise filter based on perceived intensity derived from vibration data obtained from vibration sensors.

First, a filter is created that divides the 100-20000 Hz band into certain frequency range. In this paper, filter is divided into 20 Hz. The intensity is calculated for each frequency of the base signal using Empirical Mode Decomposition (EMD), and the amount of ego noise removal is updated as the filter value as shown in Fig.1. Every 2.5 ms, the perceptual intensity for each frequency band is compared with the corresponding filter value, and only when the input intensity is greater than the filter value, the difference is added to the filter to update the noise removal amount.

Occasionally, an intensity significantly exceeding the Intensity of the target ego-noise is measured, resulting in excessive noise reduction. To prevent this, a coefficient is set to dynamically adjust the additive ratio to the difference value. The amount added when updating this intensity is shown in the following equation. Here, $F_n(f)$ is a value of filer at time n in frequency range f and is a value of filer at time n+1 in frequency range f, and is an input intensity. This process is repeated in real-time while presenting vibrations, and the measurement of noise ends when sufficient noise reduction is achieved. .

$$F_{n+1}(f) = F_n(f) + (I_{in}(f) - F_n(f))\left(\frac{F_n(f)}{I_{in}(f)}\right)^2 \quad (2)$$

## 3 SYSTEM CONFIGURATION

To present vibrations, amplitude-modulated waves were generated from vibration signals obtained by a vibration sensor. System configuration is shown in Fig.2. The vibration signals will be converted to digital signals using a USB audio interface (Zoom, AMS-24), processed on a PC with noise subtraction as shown in Fig. 1, converted back to analog signals using the USB audio interface, amplified by an amplifier, and then used to actuate a vibrator.

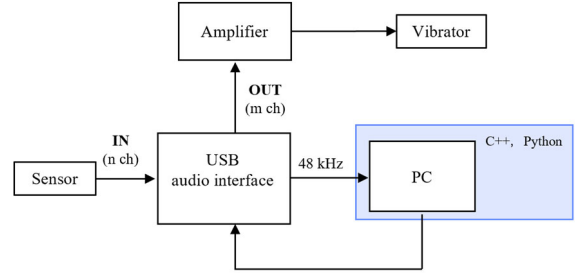

Fig.2 System Configuration

## 4 CONCLUSION

Using these two methods, we developed noise spectral subtraction method that incorporates human perception. This method effectively eliminates ego noise generated by the robot's own motion. In our demonstration, you can experience how noise is effectively reduced using a leader and follower robot.